\documentclass[letterpaper, 10 pt, conference]{ieeeconf}  

\IEEEoverridecommandlockouts                              

\usepackage{graphics} 
\usepackage{amsmath} 
\usepackage{amssymb}  
\usepackage{graphicx}
\usepackage{subcaption}
\usepackage{algorithm}
\usepackage{caption}
\usepackage{booktabs}
\usepackage{multirow}
\usepackage{url}
\usepackage{algpseudocode}
\usepackage{xcolor}

\newcommand{\ProjectName}{COMRES-VLM}
\usepackage{todonotes}

\title{\LARGE \bf
\ProjectName: Coordinated Multi-Robot Exploration and Search using Vision Language Models
}

\author{Ruiyang Wang, Hao-Lun Hsu, David Hunt, Jiwoo Kim, Shaocheng Luo, and Miroslav Pajic} 

\begin{document}

\maketitle
\thispagestyle{empty}
\pagestyle{empty}

\begin{abstract}
Autonomous exploration and object search in unknown indoor environments remain challenging for multi-robot systems (MRS). Traditional approaches often rely on greedy frontier assignment strategies with limited inter-robot coordination. In this work, we present Coordinated Multi-Robot Exploration and Search using Vision Language Models (COMRES-VLM), a novel framework that leverages Vision Language Models (VLMs) for intelligent coordination of MRS tasked with efficient exploration and target object search. COMRES-VLM integrates real-time frontier cluster extraction and topological skeleton analysis with VLM reasoning over shared occupancy maps, robot states, and optional natural language priors, in order to generate globally consistent waypoint assignments. Extensive experiments in large-scale simulated indoor environments with up to six robots demonstrate that COMRES-VLM consistently outperforms state-of-the-art coordination methods, including Capacitated Vehicle Routing Problem (CVRP) and Voronoi-based planners, achieving 10.2\% faster exploration completion and 55.7\% higher object search efficiency. Notably, COMRES-VLM enables natural language-based object search capabilities, allowing human operators to provide high-level semantic guidance that traditional algorithms cannot interpret. 

\end{abstract}

\begingroup
\addtocounter{footnote}{-1}%
\endgroup

\section{Introduction}
Coordinated exploration and object search remains a fundamental challenge for multi-robot systems (MRS), particularly in unknown and dynamic environments such as disaster zones, industrial facilities, and subterranean caves~\cite{sugiyama2013real, dunbabin2012robots, rouvcek2019darpa}. Achieving efficient, scalable exploration with MRS requires sophisticated planning strategies that effectively balance local sensing, task allocation, and global coordination.

Classical sampling-based planners, such as Rapidly-exploring Random Trees (RRT)~\cite{bircher2016receding}, have been extensively used in robotic exploration. While effective for rapid deployment, they often result in redundant or inefficient trajectories, particularly in cluttered or expansive environments~\cite{wittinghistory, selin2019efficient}. Frontier-based exploration has emerged as a more information-driven alternative, guiding robots toward the boundary between known and adjacent unknown regions to maximize new information gain. However, most implementations rely on greedy heuristic assignments, typically selecting frontiers based solely on proximity or estimated local utility, without considering global coordination or workload balance. 

\begin{figure}
    \centering
    \includegraphics[width=0.72\linewidth]{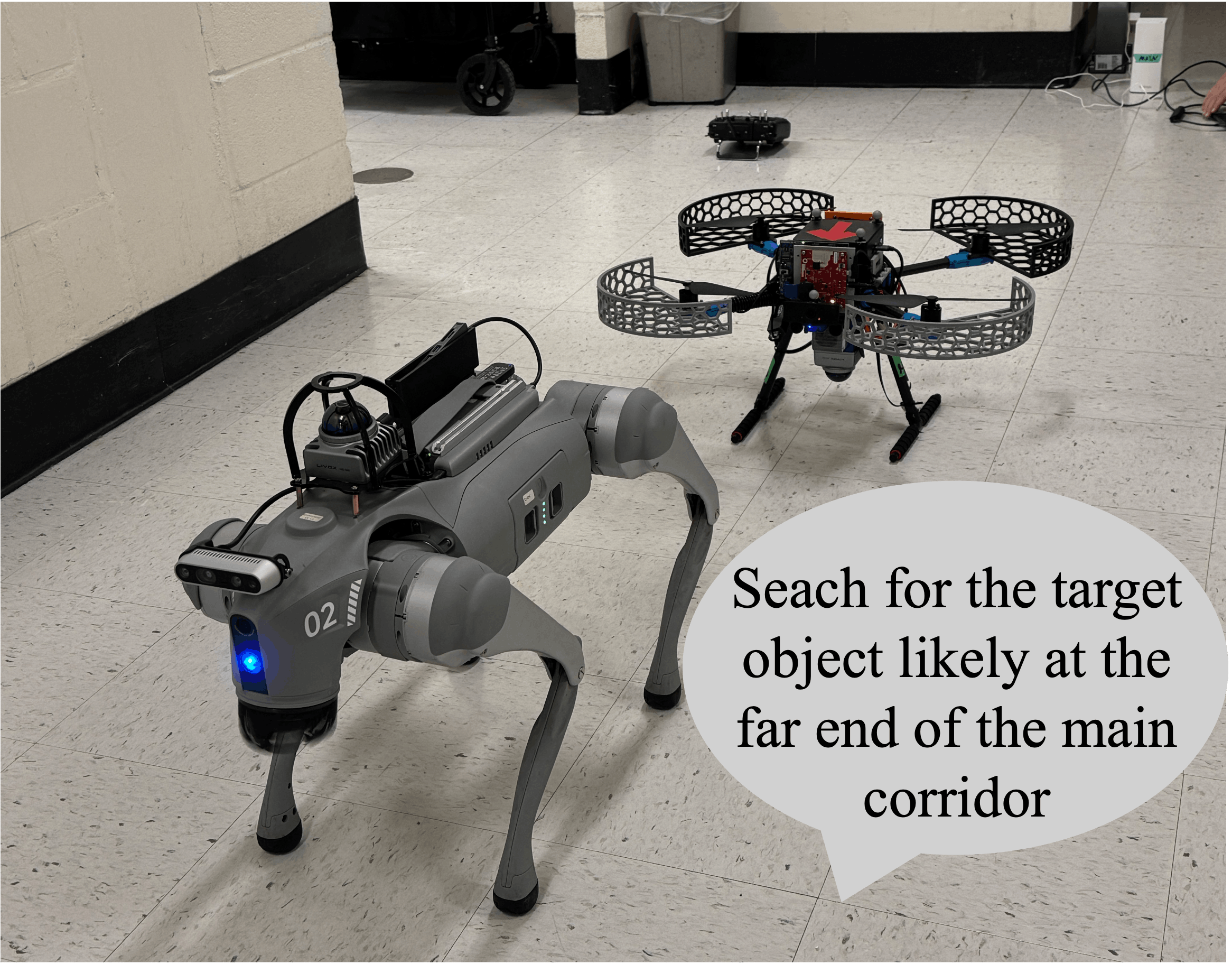}
    \caption{MRS deployed for indoor search with unstructured language prior information from human.}
    \label{fig:mrs}
    \vspace{-4mm}
\end{figure}

Recent advances in Large Language Models (LLMs) and Vision-Language Models (VLMs) offer promising avenues for addressing these limitations, owing to their capabilities in reasoning and high-level decision-making across robotics applications~\cite{chen2024scalable, kannan2024smart, calandra2018more}. Foundation-model-based systems such as VIMA~\cite{jiang2023vima} and RT-2~\cite{zitkovich2023rt} demonstrate that large-scale multimodal models can generalize structured reasoning across perception and action domains. Such capabilities make VLMs well-suited for multi-robot coordination tasks, which require both spatial efficiency and semantic interpretation that traditional geometric or heuristic planners struggle to integrate. As illustrated in Fig.~\ref{fig:mrs}, high-level natural language instructions (e.g., \textit{“search for the target object likely at the far end of the main corridor”}) can be grounded in spatial structure to guide coordinated robot behavior. Recent works such as CoNavGPT~\cite{yu2023co} and MCoCoNav~\cite{shen2025enhancing} extend LLM/VLM reasoning to multi-robot coordination with semantic inputs. However, their methods require $\mathcal{O}(n)$ sequential queries to assign $n$ waypoints, each processing $\mathcal{O}(m)$ frontier clusters. This results in an overall $\mathcal{O}(nm)$ structure incurs significant computational overhead, limiting scalability and real-time multi-robot coordination. These observations suggest the need for a coordination paradigm that preserves semantic reasoning while avoiding sequential per-robot inference bottlenecks.

Consequently, we introduce --- \textbf{Co}ordinated \textbf{M}ulti-\textbf{R}obot \textbf{E}xploration and \textbf{S}earch using \textbf{V}ision \textbf{L}anguage \textbf{M}odels (\ProjectName), a novel framework that leverages a pre-trained VLM as a centralized high-level planner for efficient multi-robot exploration and object search in unknown environments. \ProjectName~integrates both structured spatial information (e.g., extracted frontiers and topological skeletons from an occupancy map) and unstructured semantic cues (e.g., natural language hints) to generate meaningful waypoint sequences for each robot.~Then, \ProjectName~processes the global occupancy map as a single grayscale image input with structured spatial information~overlays to the VLM, enabling visual spatial~reasoning in near real-time.

A key advantage of \ProjectName~is its ability to incorporate semantic guidance. For instance, when given an instruction grounded in spatial structure to guide coordinated robot behavior, the VLM can jointly reason over the spatial layout and instruction to prioritize relevant areas. This enables context-aware and adaptive planning that significantly extends beyond the capabilities of traditional heuristic or geometry-driven approaches, effectively bridging robotic perception and human-like semantic reasoning.

Finally, we evaluate \ProjectName~through extensive simulations and real-world experiments across varying team sizes and diverse environment layouts. 
In summary, the main contributions of this work are:
\begin{itemize}

\item We propose \ProjectName, a centralized approach that leverages a VLM foundation model to reason over structured geometric and unstructured semantic input, and assign informative waypoints to each robot.

\item We evaluate \ProjectName~against two SOTA MRS coordination approaches, including \textit{RACER} \cite{zhou2023racer} and \textit{AEP} \cite{selin2019efficient} with Dynamic Voronoi Cell (DVC) \cite{cortes2004coverage}, showing its effectiveness in MRS exploration and search.

\item We demonstrate that \ProjectName~uniquely supports language-based object search, significantly outperforming traditional strategies when high-level prior knowledge is available. Our method achieved up to $\mathbf{55.7\%}$ reduction in search completion timestep. We further validate the proposed framework through real-world multi-robot experiments.

\end{itemize}
\section{Related Work} 
Autonomous robotic exploration has been studied extensively. Here, we review representative works in three key relevant areas:
\textit{(1)}~classical exploration strategies, \textit{(2)}~multi-robot coordination mechanisms, and \textit{(3)}~the emerging integration of VLMs in robotic planning.

\subsubsection{Classical Exploration Strategies}
Traditional robotics exploration methods 
can be broadly categorized into sampling-based planning and frontier-based exploration. 
Sampling-based approaches (e.g.,~\cite{papachristos2017uncertainty,bircher2018receding}) generate candidate viewpoints in the vicinity of the robot and rank them using a utility function that balances expected information gain and travel cost. While effective for local optimization and responsive to immediate perceptual feedback, these methods typically commit to short-horizon decisions based on locally sampled viewpoints. As a result, they often lack global coordination and long-horizon reasoning, limiting their effectiveness in MRS.

Frontier-based exploration, in contrast, explicitly targets the boundary between known free space and adjacent unknown regions \cite{zhou2021fuel,mobarhani2011histogram}. By driving robots toward these frontiers, the method naturally encourages outward expansion of explored space and provides stronger long-horizon guidance. To improve robustness and reduce noise from raw frontier cells, clustering techniques, such as Mean-Shift~\cite{caiza2024autonomous} grouping, are commonly employed to aggregate frontier cells into representative regions. This structured frontier representation improves stability and computational efficiency, particularly in multi-robot settings.

Hybrid approaches attempt to combine the complementary strengths of both paradigms by integrating global frontier selection with local Next-Best-View (NBV) refinement~\cite{charrow2015information,dai2020fast}. These methods aim to direct robots toward globally informative regions while allowing fine-grained viewpoint optimization locally. Although such strategies improve exploration efficiency and reduce myopic behavior, they remain fundamentally grounded in structured geometric representations derived from occupancy grids or point clouds. Consequently, decision making is restricted to spatial heuristics and predefined utility formulations, without the capability to incorporate or adapt to unstructured high-level inputs, such as natural language descriptions or task priors.

\subsubsection{Multi-Robot Exploration and Coordination}
While MRS can significantly accelerate exploration by distributing workload among robots, coordinating agents in partially known environments remains challenging. A common approach is greedy task assignment~\cite{tran2022frontier, butzke2011planning, burgard2005coordinated} that allocates frontiers or viewpoints based on robot proximity and individual utility estimates. These methods are computationally efficient but often result in unbalanced and inefficient multi-robot exploration due to a lack of global reasoning.

Decentralized coordination mechanisms, including auction-based task allocation~\cite{zlot2002multi, smith2018distributed}, potential field methods \cite{yu2021smmr}, and Monte Carlo Tree Search (MCTS) strategies \cite{bone2023decentralised}, enable scalable decision-making without requiring full global synchronization. However, because each robot operates with limited access to global information, these approaches often exhibit reduced global optimality, particularly in large-scale or complicated environments where coordinated long-horizon planning is critical.

More advanced coordination strategies formulate multi-robot exploration as a combinatorial optimization problem, such as a Capacitated Vehicle Routing Problem (CVRP) \cite{zhou2023racer}, or Dynamic Voronoi Cell (DVC) partitioning \cite{wang2024sensor,hu2020voronoi}, to spatially decompose the environment among robots. By explicitly modeling workload balancing and spatial distribution, these centralized approaches typically achieve strong coverage efficiency and serve as competitive baselines for the proposed \ProjectName~framework.

\subsubsection{Language Models in Robotics Planning}
Recent LLMs and VLMs advances 
have sparked interest in using language-guided reasoning for 
MRS collaborations. Early 
focus was on task decomposition and role assignment using structured language prompts \cite{yu2024mhrc, mandi2024roco, rajvanshi2025sayconav}, often assuming known environments and fixed atomic actions. Moreover, inter-robot coordination 
has relied on extensive rounds of LLM-based dialog or distributed reasoning, resulting in significant computational overhead that hinders real-time performance and scalability. VLMs were also used for object search in unknown settings under homogeneous MRS~\cite{yu2023co, shen2025enhancing}, but they largely depended on local visual observations (e.g., camera inputs) to make short-horizon decisions, typically selecting a single promising frontier for each robot without considering long-horizon planning and coordination. In particular, MCoCoNav~\cite{shen2025enhancing} processes four VLM modules and an image for each frontier cluster to allocate individual robots, creating significant computational overhead that constrains real-time deployment.

To the best of our knowledge, following~\cite{yu2023co, shen2025enhancing}, \ProjectName~is the first centralized framework to combine structured spatial representations, such as frontier clusters and topological skeletons extracted from occupancy maps, with natural language cues for long-horizon multi-robot coordination in unknown environments. Rather than replacing geometric planning, \ProjectName~augments it with high-level reasoning over spatial structure and task context, offering a unified approach to explore and search in partially observed settings.

\section{Problem Definition}
We consider 
an MRS consisting of $M$ robots. 
The state of robot $i$ at timestep $t$ is defined by its 2D position vector $X^t_i = [x^t_i \; y^t_i]^T \in \mathbb{R}^2$. To ensure collision avoidance, robots must maintain a minimum safety distance $d_{safe} > 0$ from one another, such that $||X^t_i - X^t_j||_2 \geq d_{safe}$ for all $i \neq j$ and all timesteps $t$. Each robot $i$ is equipped with a 2D sensor characterized by a maximum detection range $d^i_{det} \in \mathbb{R}$ and an angular range $\theta^i_{det}$ field of view (FoV). Also, each robot $i$ has a maximum velocity $V^i_{max}$ such that $||X^{t+1}_i - X^{t}_i||_2 \leq ||V^i_{max}||_2 \;$, for all $t$. At any given timestep, the robot can observe all grid cells within its FoV if those cells are not occluded by obstacles. 

The 
MRS objective is to either \emph{explore} an indoor and previously unknown 2D space $\mathcal{G} \in \mathbb{R}^2$ or to \emph{search} for a target object located at an unknown position $g_{int} \in \mathcal{G}$. The environment is represented as a grid map $\mathbf{G} \in \mathbb{Z}^{H\times W}$, with a fixed resolution $r$, height $H \in \mathbb{R}$ and width $W \in \mathbb{R}$. Each cell $\mathbf{g} \in \mathbf{G}$ is labeled as \textit{unknown}, \textit{free}, or \textit{occupied}, which we visualize as \textit{white}, \textit{gray}, and \textit{black} accordingly. These classifications induce a partition of the environment into three disjoint subspaces: the free space  $\mathcal{G}_{fr}$, the occupied space $\mathcal{G}_{oc}$, and the unknown space $\mathcal{G}_{un}$, such that at all timesteps $\mathcal{G} = \mathcal{G}_{fr} \cup \mathcal{G}_{oc} \cup \mathcal{G}_{un}$.
The exploration problem is considered fully solved when $\pi_{threshold} \in [0, 1]$ of the full observable regions have been explored as $|\mathcal{G}_{oc} \cup \mathcal{G}_{fr}| \geq \pi_{threshold}|\mathcal{G} \setminus \mathcal{G}_{res}|$, where $\mathcal{G}_{res}$ denotes permanently inaccessible or occluded areas. In the case of a search task, the problem is considered solved once the object of interest has been observed such that $g_{int} \in \mathcal{G}_{fr} \cup \mathcal{G}_{oc}$.

\section{\ProjectName~Framework}
\begin{figure}[!t]
    \centering
    \includegraphics[width=1.0\linewidth]{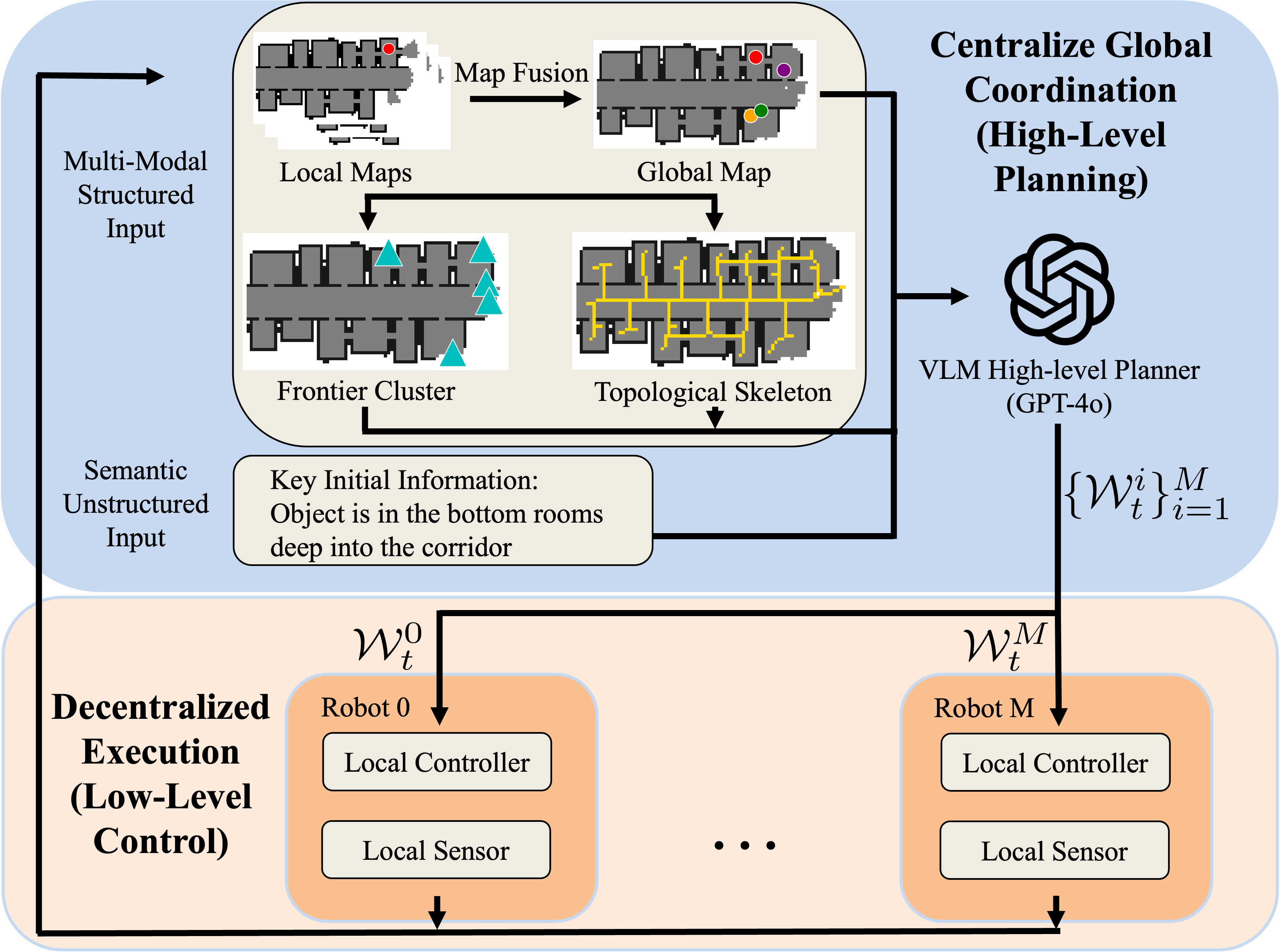}
    \caption{\small{\ProjectName~Planning Pipeline. At coordination time, the robot's local maps with their location shown in circles are aggregated into the global map. Then the frontier clusters (shown in blue triangles) and topological skeletons (shown in yellow lines) are extracted and forwarded with the semantic unstructured input in to the VLM high-level planner for waypoint assignments globally. Each robot receives the assigned waypoint and execute locally.}}
    \label{fig:vlm_pipeline}
    \vspace{-4mm}
\end{figure}

The \ProjectName~framework adopts a centralized coordination and decentralized execution paradigm (Fig.~\ref{fig:vlm_pipeline}). At each coordination step, local occupancy maps from individual robots are aggregated into a global representation, from which frontier clusters and topological skeleton structures are extracted. Also, high-level semantic cues expressed in natural language are incorporated as auxiliary guidance. By fusing these complementary sources of information, \ProjectName~enables more informed and adaptive planning, allowing robot teams to coordinate effectively and explore previously unknown environments with~greater~efficiency.

\subsection{Frontier Cluster Detection}
\label{sub:frontier}
Given the grid map $\mathbf{G}$, we adopt a standard frontier extraction formulation \cite{yamauchi1997frontier}. 
A grid cell $\mathbf{g} \in \mathcal{G}_{fr}$ is defined as a \emph{frontier} if it is adjacent to at least one unknown cell:
\begin{equation}
\mathbf{g} \in \mathcal{G}_{fr}
\quad \text{and} \quad
\exists \, \mathbf{g}' \in \mathcal{N}_4(\mathbf{g})
\text{ such that }
\mathbf{g}' \in \mathcal{G}_{un},
\end{equation}
where $\mathcal{N}_4(\mathbf{g})$ denotes the 4-connected neighborhood of $\mathbf{g}$.

Frontier cells represent the boundary between explored and unexplored regions and therefore define candidate expansion locations. To obtain spatially meaningful targets, frontier cells are grouped into clusters using breadth-first search over free-space connectivity. To prevent excessive merging in large open areas, clustering is restricted by a maximum geodesic distance threshold $c_{max}$.

For each cluster, we compute its centroid and select the frontier cell closest to the centroid as its representative waypoint. Each frontier cluster is characterized by its size and representative location, and serves as a candidate region for coordinated exploration or search. Examples of frontier clusters are visualized in Fig.~\ref{fig:vlm_pipeline}. The set of frontier clusters, $\mathcal{F}$, consists of individual frontier cluster, $f_i=\{p^i_{rep},c^i_{size}\}$, with representative position $p^i_{rep}$ and cluster size $c^i_{size}$. The set $\mathcal{F}$ is then forwarded as structured input to the VLM for high-level planning. 

\subsection{Topological Skeleton Analysis}
\label{sub:skeleton}
To provide structural cues about the spatial layout of the explored region, we compute the morphological skeleton of the free space $\mathcal{G}_{fr}$. Given the grid representation $\mathbf{G} \in \mathbb{Z}^{H \times W}$, we construct a binary mask over the free cells $\mathcal{G}_{fr}$, where a grid cell $\mathbf{g} \in \mathbf{G}$ is assigned value $1$ if $\mathbf{g} \in \mathcal{G}_{fr}$ and $0$ otherwise. 

We apply the Zhang–Suen thinning algorithm~\cite{zhang1984fast}, an iterative parallel procedure that progressively deletes boundary free cells from the occupancy grid while preserving connectivity and topology. Specifically, at each iteration, free cells located on the outer boundary of the free-space region are considered for removal. A boundary free cell is deleted only if it satisfies three conditions based on its 8-neighborhood within $\mathbf{G}$: (1) it has between 2 and 6 neighboring free cells, (2) its ordered neighborhood contains exactly one 0-to-1 transition to ensure it lies on a contour, and (3) designated neighbor subsets contain at least one non-free cell to prevent disconnection. Deletions are carried out in two alternating sub-iterations to avoid directional bias and preserve endpoints. The process repeats until no further boundary cells can be removed. The resulting skeleton $\mathcal{S} \subset \mathcal{G}_{fr}$ is a one-cell-wide structure that retains the connectivity of the free space while suppressing redundant geometric thickness.
We overlay the extracted skeleton $\mathcal{S}$ onto the visual map provided to the VLM. This topological abstraction enhances the VLM’s perception of global connectivity and branching structure, facilitating more informed coordination and waypoint allocation decisions.

\subsection{VLM Coordinated Explore and Search}
Unlike traditional methods that rely exclusively on frontiers \cite{caiza2024autonomous} or information gains \cite{selin2019efficient}, \ProjectName~enables the VLM to jointly reason over multiple information sources, including both frontiers clusters and topological skeletons with multiple modalities, such as a shared occupancy map in image and high-level natural language initial information. These multimodal inputs are aggregated into a unified input representation $\mathcal{Z}_t$ and passed to the VLM via a structured prompt template. The centralized coordination is performed periodically with a fixed replanning horizon $T_H \in \mathbb{N}$. The VLM outputs a sequence of waypoints $\{\mathcal{W}_i^t\}_{i=1}^M$ for each robot. This allows \ProjectName~to act as a centralized high-level planner, assigning waypoints that are grounded in both map geometry and task context. 

In this paper, we used OpenAI GPT-4o as the VLM foundation model. An overview for the \ProjectName~algorithm is provided in Algorithm.~\ref{alg:\ProjectName}. At each coordination step $t$ satisfying $t_{\text{since\_plan}} \geq T_H$, local maps from all robots are fused into a global occupancy map $\mathbf{G}_t$. Frontier clustering (Sec.~\ref{sub:frontier}) and skeleton extraction (Sec.~\ref{sub:skeleton}) are then performed~on~this~map. The resulting frontier clusters $\mathcal{F}_t$, skeleton structure $\mathcal{S}_t$ at timestep $t$, and robot positions $\{\mathbf{x}_i^t\}_{i=1}^M$ constitute the structured spatial input. These elements are rendered onto the global map, as shown in Fig.~\ref{fig:vlm_pipeline}, and provided to the VLM as visual input. In addition, the frontier cluster information and robot positions are explicitly encoded as textual descriptions and supplied alongside the visual input to ensure precise spatial grounding.

\begin{algorithm}[t!]
\caption{\ProjectName}

\label{alg:\ProjectName}
\begin{algorithmic}[1]

\State $t \gets 0$, $t_{\text{since\_plan}} \gets 0$
\State \textbf{Optional:} $\mathcal{I}_{\text{init}}$ Key Initial Information

\While{task not complete}

    \If{$t_{\text{since\_plan}} \geq T_H$}
    
        \State Fuse local maps $\rightarrow \mathbf{G}_t$
        \State Extract frontiers $\mathcal{F}_t$  and skeleton $\mathcal{S}_t$
        \State $\mathcal{Z}_t \gets (\mathbf{G}_t, \mathcal{F}_t, \mathcal{S}_t, \{\mathbf{x}_i^t\}, \mathcal{I}_{\text{init}})$
        \State $\{\mathcal{W}_i^t\} \gets \text{VLM}(\mathcal{Z}_t)$
        
        \State $t_{\text{since\_plan}} \gets 0$
        
    \EndIf
    
    \State Robots execute $\{\mathcal{W}_i^t\}_{i=1}^M$ locally
    \State $t \gets t + 1$, $t_{\text{since\_plan}} \gets t_{\text{since\_plan}} + 1$
    
\EndWhile
\end{algorithmic}
\end{algorithm}

\begin{figure}
    \centering
    \includegraphics[width=1.0\linewidth]{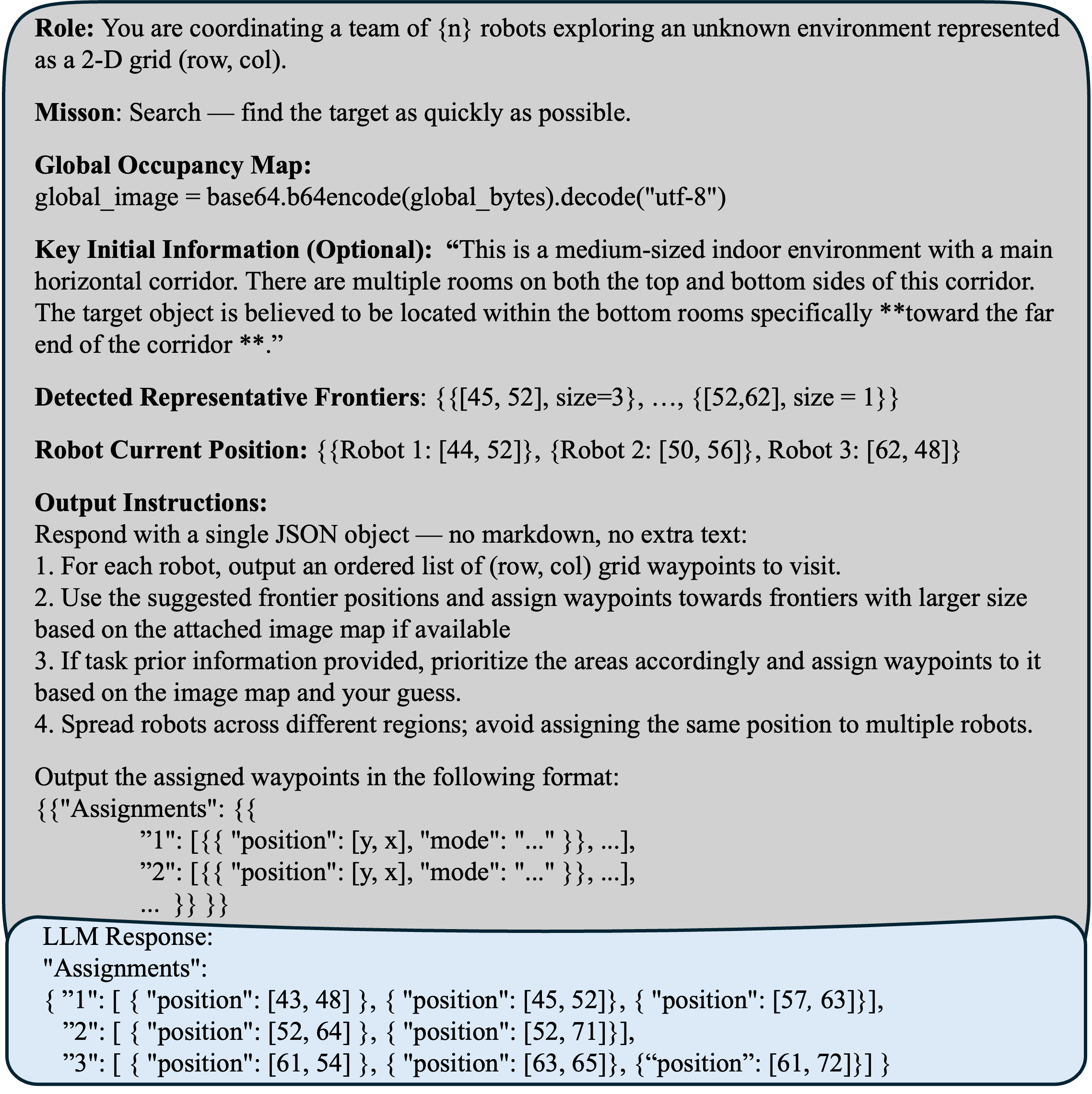}
    \caption{Example \ProjectName~query and response.}
    \label{fig:prompt}
    \vspace{-2mm}
\end{figure}

The \emph{unstructured input} consists of textual information including the task prompt for exploration or search, and optional natural language priors, referred to as the \emph{Key Initial Information}. Such priors may provide high-level semantic guidance, for example, a statement such as \textit{the object of interest is likely located in deep into corridor and in the bottom rooms}. The VLM integrates this semantic information with the structured spatial inputs. This semantic grounding enables the planner to go beyond purely geometry-driven decision making and adapt its strategy based on abstract goals or vague human specified hints. By fusing this key initial information with real-time sensory and spatial data, the VLM serves as a centralized high-level planner that generates context-aware, goal-driven waypoint assignments. An example of a VLM query and response is shown in Fig.~\ref{fig:prompt}.

To mitigate VLM hallucinations and formatting inconsistencies, each coordination query is retried up to three times if a valid JSON response is not returned. The local controller further rejects infeasible or unreachable waypoints using the current local map. If all assigned waypoints are completed or invalid, the robot reverts to a greedy policy that navigates to the nearest frontier cluster. This hierarchical safeguard ensures robustness and prevents idle behavior; in practice, re-querying or fallback occurs in \emph{less than 1\% of total runs}.

\subsection{Design Insights}
Classical multi-robot exploration strategies, such as distance-based frontier assignment, Voronoi partitioning, or combinatorial routing (e.g., CVRP), while effective for workload balancing and coverage efficiency, these approaches fundamentally operate on predefined utility formulations and local spatial metrics. Consequently, they lack mechanisms to incorporate high-level task intent, structural prioritization, or ambiguous semantic guidance.

\ProjectName~addresses this limitation by elevating coordination from local utility optimization to global structural reasoning. The VLM receives the global occupancy grid as a visual input to ensure precise spatial grounding, while additional structured annotations, including frontier clusters, topological skeleton overlays, and robot states, provide explicit structural cues about connectivity and region organization. Rather than relying exclusively on raw geometric proximity or handcrafted utility functions, the VLM reasons jointly over this layered representation, which preserves global spatial layout while emphasizing corridors, room groupings, and candidate expansion regions.

Importantly, \ProjectName~preserves classical robustness by separating high-level reasoning from low-level execution. The VLM generates waypoint sequences at a fixed replanning horizon $T_H$, while robots execute them using conventional planners with safety validation and fallback mechanisms. Thus, the foundation model functions as a centralized structural coordination layer built atop reliable geometric control, rather than replacing it. From this perspective, \ProjectName~can be interpreted as learning an implicit global coordination heuristic in zero shot conditioned on environment topology and task priors, extending beyond handcrafted geometric formulations.

\section{Simulation Results}
\begin{figure}[!t]
    \centering
    \includegraphics[width=0.668\linewidth]{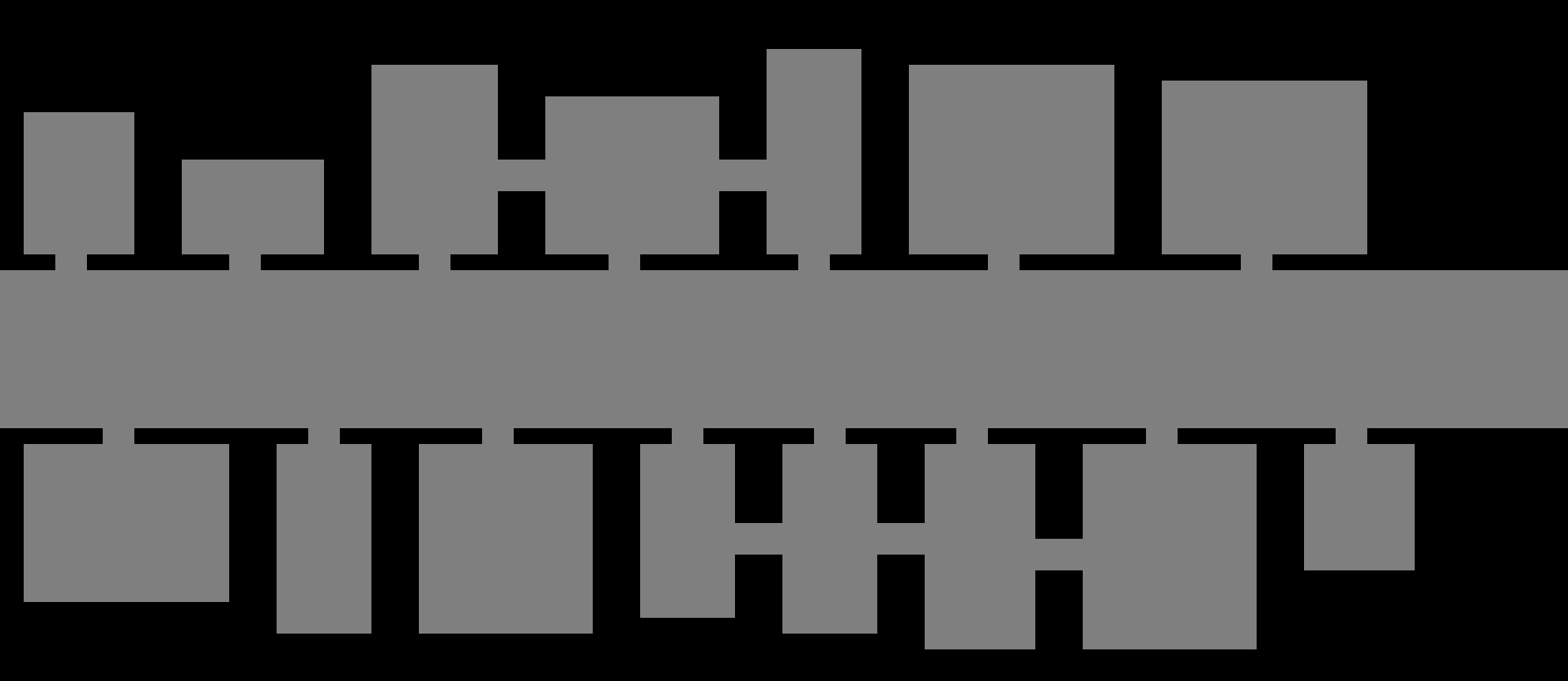}
    \caption{Example simulation environments used for evaluation.}
    \label{fig:medium_map}
    \vspace{-5mm}
\end{figure}

We evaluate the effectiveness of \ProjectName~framework in  comprehensive simulations for both exploration and search tasks in structured environments, as illustrated in Fig.~\ref{fig:medium_map}. Each robot is equipped with a low-level A* path planner for navigating toward its assigned waypoints. To enable decentralized collision avoidance, each robot maintains a temporary local map with the current positions of other robots modeled as circular obstacles with a $d_{safe} = 1$ radius.

\subsection{Baselines}
We compared {\ProjectName}'s performance to several baselines:
\textit{(i) RACER}~\cite{zhou2023racer} is a leading method for coordinated multi-robot exploration. RACER decomposes the global map into hierarchical grid cells (HGrids) based on the unknown cell ratio within each region. It then performs pairwise coordination by solving a CVRP between robot pairs to allocate HGrids. After receiving their assigned HGrids, robots locally solve a \emph{Traveling Salesman Problem} (TSP) to visit frontier clusters within the first $k$ assigned HGrids. Once these frontiers are explored, each robot continues replanning locally according to its precomputed CVRP-based HGrid~ordering.

\textit{(ii) AEP}~\cite{selin2019efficient} +\textit{DVC}~\cite{cortes2004coverage}, where AEP is a state-of-the-art single-robot exploration strategy that selects waypoints by maximizing estimated utility based on expected information gain and travel cost over sampled free-space candidates. To extend AEP to the multi-robot setting, we incorporate \emph{Dynamic Voronoi Cells} (DVC) to partition the map, assigning each robot to explore its corresponding Voronoi region.

For fair comparison, we fix the global replanning horizon to $T_H = 50$ timesteps for all methods. Within each replanning interval, RACER resolves its pairwise CVRP assignments, AEP+DVC updates the Voronoi decomposition, and \ProjectName~queries GPT-4o with \texttt{max\_token} = 1024 and \texttt{temperature} = 0.2 for new waypoint assignments. Because RACER and AEP+DVC continuously generate local waypoints once a robot completes its planned trajectory, we equip \ProjectName~with a local backup controller. Specifically, if a robot finishes its assigned waypoints before the next VLM query, it temporarily navigates to the nearest frontier in its local map. This prevents robots from remaining idle while awaiting new high-level assignments, and ensures consistent utilization across all compared methods.

\subsection{Exploration and Search}
\label{simulation::1}

\begin{table*}[t]
\centering
\caption{Average completion time (in timesteps) for exploration and search tasks under different environment scales.}
\label{tab:strategy_comparison_VLM_MCES}
\small
\setlength{\tabcolsep}{3.8pt}
\begin{tabular}{lccc|ccc}
\toprule
\multirow{2}{*}{\textbf{Strategy}} 
& \multicolumn{3}{c|}{\textbf{Exploration Time}} 
& \multicolumn{3}{c}{\textbf{Search Time}} \\
\cmidrule(lr){2-4} \cmidrule(lr){5-7}
& \textbf{Small} & \textbf{Medium} & \textbf{Large}
& \textbf{Small} & \textbf{Medium} & \textbf{Large} \\
\midrule
\textit{RACER}     
& $181.25 \pm 28.70$ & $254.12 \pm 26.52$ & $368.02 \pm 33.09$
& $164.73 \pm 35.87$ & $244.73 \pm 31.89$ & $370.80 \pm 43.05$ \\

\textit{AEP+DVC}
& $171.81 \pm 22.75$ & $248.08 \pm 23.65$ & $359.85 \pm 29.15$
& $160.47 \pm 30.93$ & $244.45 \pm 27.69$ & $365.94 \pm 37.11$ \\

\textit{\ProjectName (Ours)}
& $\mathbf{156.12 \pm 25.29}$ & $ \mathbf{212.23 \pm 17.12}$ & $\mathbf{323.02 \pm 19.17}$
& $\mathbf{63.21 \pm 18.19}$ & $\mathbf{104.25 \pm 19.15}$ & $\mathbf{162.24 \pm 18.50}$ \\
\bottomrule
\end{tabular}
\vspace{-2mm}
\end{table*}

We evaluate all methods in three scales: small ($H=60, W=30$), medium ($H=120, W=60$), and large ($H=150, W=80$). Each map contains a central corridor with randomly sized rooms distributed along both sides. To avoid density saturation, we deploy 2, 4, and 6 robots in small, medium, and large maps, respectively. All robots move at unit speed and are equipped with a $d_{det}=10$ and $\theta_{det}=\pi/3$ FoV sensor.

For each scale, we generate 10 distinct environment instances, resulting in 30 unique maps. For each map, we sample 10 different initial robot configurations (and object positions for search), leading to 100 runs per scale per task. We set the coverage threshold to be $\pi_{threshold}$ = 0.95 for exploration tasks.

For pure exploration, \ProjectName~is not provided with any prior information regarding target location. All other inputs, as detailed in Fig.~\ref{fig:vlm_pipeline}, are provided for the VLM high-level planner to reason. As shown in Table.~\ref{tab:strategy_comparison_VLM_MCES}, \ProjectName~consistently outperforms both baselines across all map scales. While AEP+DVC slightly improves over RACER, \ProjectName~achieves the largest gains, with up to \textbf{14.5\% improvement} in medium-scale environments. This demonstrates that global structural reasoning over aggregated maps improves coordination efficiency even without any prior~knowledge about the system/environment.

The performance advantage becomes significantly more pronounced in search tasks. In this setting, \ProjectName~is provided with coarse prior information, such as \emph{``the object is likely located deep in the corridor and within the bottom rooms.''} This prior is converted into spatial guidance and incorporated into the high-level planning prompt. 

For this search setting, \ProjectName~is not provided with explicit frontier annotations. When strong directional priors are available, frontier overlays may inadvertently bias the VLM toward boundary-following behavior, potentially conflicting with region-level prioritization implied by the prior. This design choice is further supported by the ablation analysis in Sec.~\ref{subsec:ablation}, 
showing that removing frontier annotations improves search efficiency under informative~priors.

As shown in Table~\ref{tab:strategy_comparison_VLM_MCES}, both RACER and AEP+DVC exhibit search times comparable to their exploration times, indicating that they effectively perform near-complete environment coverage before locating the target. In contrast, \ProjectName~leverages the provided prior to bias global coordination, reducing search time to nearly half of its exploration time in large environments. The improvement reaches \textbf{55.7\%} in large maps, highlighting the benefit of integrating high-level semantic reasoning into coordinated multi-robot search.

To further assess robustness and stability under controlled conditions, we evaluate performance and computational time per replanning cycle (seconds) within a fixed testing scenario. Although stochastic components (e.g., random sampling in AEP+DVC and inference variability in \ProjectName)~remain present, the map layout, robot initial positions, and target location are kept identical across all methods to ensure a fair comparison. Each method is repeated 10 times under this same setting to capture performance variance arising purely from algorithmic stochasticity. As shown in Table~\ref{tab:fixed_scenario_comparison}, RACER exhibits zero variance in performance, as its HGrid decomposition, CVRP allocation, and TSP planning are fully deterministic under fixed inputs. In contrast, both AEP+DVC and \ProjectName~demonstrate non-zero standard deviations, reflecting their inherent stochastic components. Notably, \ProjectName~achieves exploration variance comparable to AEP+DVC, while exhibiting lower variance in search performance. This suggests that incorporating prior key initial information stabilizes target-directed coordination.

In terms of computational cost, AEP+DVC incurs the lowest overhead, since coordination only requires updating DVC partitions. RACER requires solving pairwise CVRP problems at each replanning step, resulting in higher computational time. The \ProjectName~framework incurs the largest coordination cost due to the VLM inference and waypoint generation. However, owing to the local backup controller, robots continue navigating toward nearby frontiers while awaiting new assignments, preventing idle time during inference. Thus, despite higher coordination overhead, the {\ProjectName} 
remains practical for real-time~deployment.

\begin{table}[t]
\centering
\caption{Fixed scenario performance comparison.}
\label{tab:fixed_scenario_comparison}
\small
\setlength{\tabcolsep}{2.8pt}
\begin{tabular}{lccc}
\toprule
\textbf{Method} 
& \textbf{Explore} 
& \textbf{Search} 
& \textbf{Comp. (s)} \\
\midrule
\textit{RACER}     
& $272.00 \pm 0.00$
& $272.00 \pm 0.00$
& $3.16 \pm 0.02$ \\

\textit{AEP+DVC} 
& $261.70 \pm 17.93$
& $254.60 \pm 21.89$
& $0.23 \pm 0.02$ \\

\textit{COMRES-VLM}
& $\mathbf{233.00 \pm 15.70}$
& $\mathbf{121.6 \pm 6.95}$
& $6.74\pm 4.01$ \\
\bottomrule
\end{tabular}
\end{table}

\subsection{Ablation Study}
\label{subsec:ablation}
We also investigate how each component of the VLM input contributes to the performance of \ProjectName. 
We progressively augment the input to the VLM starting from visualization alone, then adding structural skeleton information, followed by task-level prior information, and finally frontier annotations. Similar to our full-scale tests, we performed this ablation study over 10 different medium-sized maps, each with 10 different initial positions and object positions and report the mean and standard deviation of each variants over 100 runs.

\begin{table}[t]
\centering
\caption{Ablation study of the VLM input components.}
\label{tab:ablation_VLM_inputs}
\small
\setlength{\tabcolsep}{3pt}
\begin{tabular}{lcc}
\toprule
\textbf{Configuration} 
& \textbf{Explore} 
& \textbf{Search} \\
\midrule
Vision only 
& $253.64 \pm  49.71$ 
& $240.87 \pm 39.02$ \\

Vision + Skeleton 
& $244.84 \pm 17.32$ 
& $225.73 \pm 21.51$ \\

Vision + Skeleton + Prior 
& $243.05 \pm 22.87$ 
& $\mathbf{105.72 \pm 16.14}$ \\

Full ( + Frontier )
& $\mathbf{214.25 \pm 16.33}$ 
& $185.68 \pm 42.84$ \\
\bottomrule
\vspace{-8mm}
\end{tabular}
\end{table}

The results are summarized in Table~\ref{tab:ablation_VLM_inputs}. We observe that using visualization alone yields the weakest performance in both exploration and search tasks, indicating that raw spatial layout without structural cues is insufficient for effective global coordination. Incorporating the topological skeleton overlay consistently improves performance in both tasks, suggesting that explicit structural information enhances the VLM’s ability to reason about corridor connectivity and room organization. When task-level prior information is introduced, search performance improves dramatically, reducing completion time by more than 50\% compared to the vision-only baseline. This confirms that semantic guidance allows the VLM to bias waypoint assignment toward prioritized regions instead of performing uniform coverage. 

Interestingly, adding frontier annotations further improves exploration performance but degrades search performance. The reason is that in search scenarios, the combination of visualization and prior information already provides sufficient global guidance. Introducing explicit frontiers inadvertently bias the VLM toward boundary following behaviors, which conflict with targeted region prioritization. In contrast, for exploration where no prior exists, frontiers provide critical boundary awareness that enhances coverage~efficiency.

\section{Real World Experiments}
\begin{figure}[!t]
    \centering
    \subcaptionbox{Robots Setup\label{fig:robot_setup}}[0.51\linewidth]{
        \includegraphics[width=\linewidth]{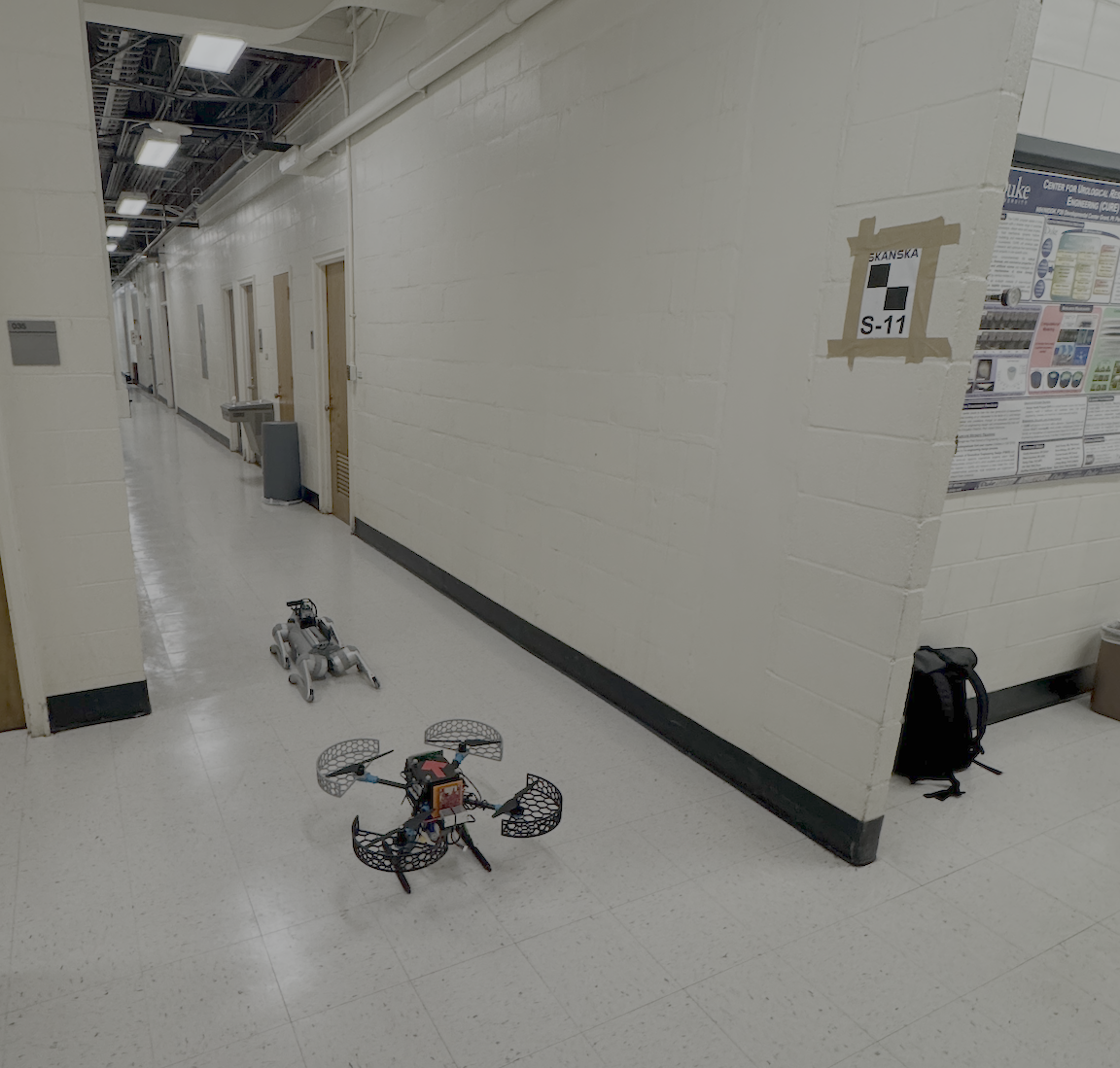}
    }
    \hspace{1mm}
    \subcaptionbox{Ground Truth Map\label{fig:ground_truth_map}}[0.434\linewidth]{
        \includegraphics[width=\linewidth]{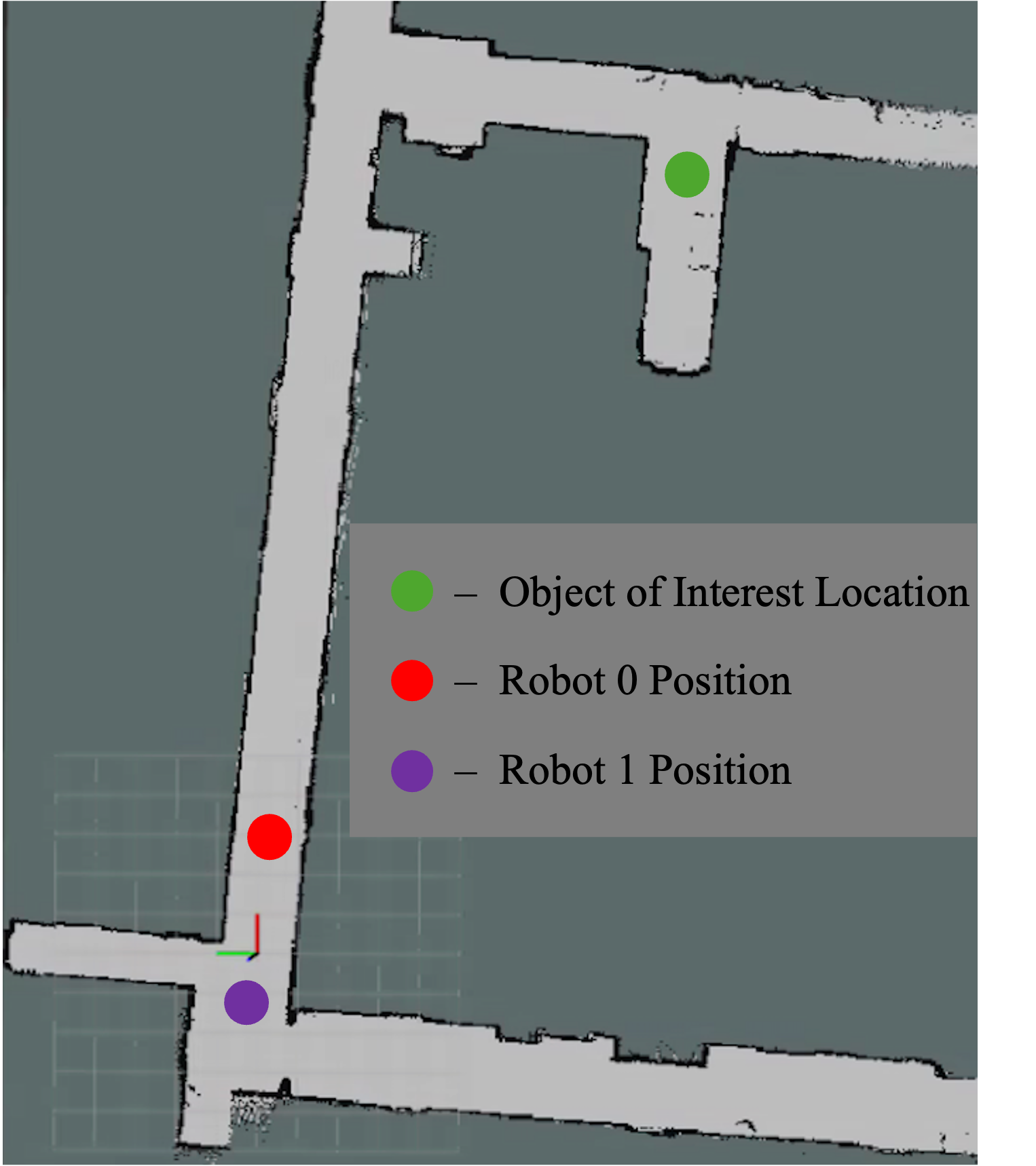}
    }
    \vspace{-2pt}
    \caption{Search experiment conducted in a hall}
    \label{fig:real_env}
    \vspace{-6mm}
\end{figure}

To evaluate the applicability of the \ProjectName~framework in coordinating MRS with different locomotion abilities in realistic settings, we conducted experiments in indoor building environments that resemble the scenarios used in simulation. The team consisted of a Unitree Go2 quadruped and a customized X500 quadrotor equipped with an Intel NUC (11th Gen i7 CPU) (Fig.~\ref{fig:robot_setup}). Both robots carried Livox Mid-360 LiDAR sensors for localization and mapping. For safety, the drone’s maximum velocity was limited to 1.0 m/s, while the Go2 operated up to 2.5 m/s.

The robots (red: Go2, purple: drone) were tasked with searching for a target (green point in Fig.~\ref{fig:ground_truth_map}), given prior knowledge that it was located deep in the corridor to the right. Each robot performed localization with SLAM Toolbox~\cite{macenski2021slam} and local waypoint following with Nav2~\cite{macenski2020marathon} on onboard computers. Map and pose updates were transmitted via Wi-Fi to a central computer (Intel i7 CPU, 32 GB RAM), which executed the VLM-based planner and assigned waypoints in near real time with a replanning horizon of 20 seconds. The experiments demonstrate that \ProjectName~can coordinate an MRS in realistic indoor environments, achieving near real-time planning and execution (a video is attached as supplementary material).

\section{Discussion and Future Directions}
The results demonstrate that \ProjectName~improves exploration efficiency even without semantic priors. Unlike RACER and AEP+DVC, which rely on locally defined geometric objectives or spatial partitions, \ProjectName~reasons over a unified global structural representation that integrates the occupancy map, frontier clusters, and topological skeleton. This enables implicit long-horizon coordination, promotes early symmetry breaking in corridor-like environments, reduces redundant frontier expansion, and improves workload balance without explicitly solving combinatorial allocation problems. When directional priors are available, the VLM biases global waypoint assignment toward likely target regions, shifting from uniform coverage to prioritized search and significantly reducing search time. Compared to prior VLM-based frameworks such as CoNavGPT~\cite{yu2023co} and MCoCoNav~\cite{shen2025enhancing}, which evaluate frontier clusters and assign a single waypoint per robot using separate visual inputs, \ProjectName~uses a single aggregated visual input and performs one global reasoning step to generate waypoint sequences for all robots, enabling coordinated long-horizon planning without sequential frontier inference.

Despite these gains, centralized VLM-based coordination faces scalability challenges as team size increases, since the joint decision space and prompt complexity grow. A natural extension of \ProjectName~is a hierarchical coordination architecture. In such a framework, in which a higher-level VLM agent would allocate subregions or macro-objectives. Lower-level VLM agents, each responsible for a subset of robots, would then perform fine-grained waypoint assignment within their designated regions. Such hierarchical design bounds prompt size, reduces reasoning dimensionality, and aligns with the decomposition of large environments.

Another promising direction involves integrating richer~onboard sensing modalities, such as cameras, to enable semantic perception beyond static priors. In the current framework, task-level guidance is provided as coarse initial information. Incorporating visual semantic cues extracted online, such as object categories or room types, would allow the foundation models to adapt dynamically as new semantic evidence emerges. This would shift coordination from prior-guided reasoning to perception-driven semantic exploration, further bridging geometric planning and high-level reasoning.

Together, hierarchical VLM coordination and semantic perception integration represent key steps toward scalable, adaptive, and context-aware MRS capable of operating in large, partially observed environments.

\section{Conclusion}
In this paper, we presented \ProjectName, a framework that introduces VLMs as a centralized high-level coordination layer for multi-robot exploration and search in unknown environments. By reasoning jointly over structured spatial abstractions and task-level language input, the proposed system enables context-aware waypoint allocation beyond purely geometry-driven formulations. Extensive simulation and real-world experiments validate that this approach improves exploration efficiency and significantly accelerates target-directed search when semantic priors are available.

Beyond empirical gains, this work highlights a broader perspective: multi-robot coordination need not be restricted to explicitly defined geometric utility functions. Instead, high-level reasoning models can operate over abstracted spatial representations to produce global coordination strategies while preserving classical low-level control and safety mechanisms. This separation between global reasoning and local execution provides a practical pathway for integrating foundation models into real robotic systems.

Looking forward, scaling VLM-based coordination to larger teams motivates hierarchical reasoning architectures. A layered design in which higher-level VLM agents allocate coarse spatial objectives and lower-level agents manage local coordination could bound reasoning complexity while maintaining global awareness. Additionally, integrating richer onboard sensing modalities, such as camera-based semantic perception, would enable adaptive planning driven by real-time environmental cues rather than static priors alone.

We believe \ProjectName~represents a step toward MRS that combine structured geometric planning with high-level reasoning, paving the way for semantically informed coordination in complex, partially observed environments.

\bibliographystyle{IEEEtran}
\bibliography{bib}
\end{document}